%% file: guided_inpainting_paper.tex
\documentclass[10pt,twocolumn,letterpaper]{article}
\usepackage{cvpr}
\usepackage{graphicx}
\usepackage{amsmath,amssymb} 
\usepackage{color}
\usepackage{algorithm}
\usepackage[noend]{algpseudocode}
\usepackage{multirow}
\usepackage{float}
\usepackage{longtable}
\usepackage[T1]{fontenc}
\usepackage{epsfig}
\usepackage{authblk}
\usepackage[symbol*]{footmisc}

\DefineFNsymbolsTM{otherfnsymbols}{%
  \textexclamdown \wr
  \textsection   \mathsection
  \textdagger    \dagger
  \textdaggerdbl \ddagger
  \textasteriskcentered *
  \textbardbl    \|%
  \textparagraph \mathparagraph
}%

\setfnsymbol{otherfnsymbols}

\makeatletter
\renewcommand\AB@affilsepx{, \protect\Affilfont}
\makeatother
\cvprfinalcopy

\ifcvprfinal\fi
\begin{document}

\title{Image Inpainting using Block-wise Procedural Training with Annealed Adversarial Counterpart} 

\author[1]{Chao Yang}
\author[1]{Yuhang Song}
\author[2]{Xiaofeng Liu}
\author[3]{Qingming Tang}
\author[1]{C.-C. Jay Kuo}
\affil[1]{USC}
\affil[2]{Carnegie Mellon University}
\affil[3]{Toyota Technological Institute at Chicago}
\affil[ ]{\small \texttt{\{chaoy,yuhangso\}@usc.edu, liuxiaofeng@cmu.edu,  qmtang@ttic.edu, cckuo@sipi.usc.edu}}

\maketitle

\input{0_Abstract.tex}

\input{1_Introduction.tex}


\input{3_Approach.tex}

\input{4_Experiments.tex}

\input{5_Conclusions.tex}


\bibliographystyle{ieee}
\bibliography{egbib}
\end{document}

%% file: 0_Abstract.tex
\begin{abstract}
Recent advances in deep generative models have shown promising potential in image inpanting, which refers to the task of predicting missing pixel values of an incomplete image using the known context. However, existing methods can be slow or generate unsatisfying results with easily detectable flaws. In addition, there is often perceivable discontinuity near the holes and require further post-processing to blend the results. We present a new approach to address the difficulty of training a very deep generative model to synthesize high-quality photo-realistic inpainting. Our model uses conditional generative adversarial networks (conditional GANs) as the backbone, and we introduce a novel block-wise procedural training scheme to stabilize the training while we increase the network depth. We also propose a new strategy called adversarial loss annealing to reduce the artifacts. We further describe several losses specifically designed for inpainting and show their effectiveness. Extensive experiments and user-study show that our approach outperforms existing methods in several tasks such as inpainting, face completion and image harmonization. Finally, we show our framework can be easily used as a tool for interactive guided inpainting, demonstrating its practical value to solve common real-world challenges. 
\end{abstract}

%% file: 1_Introduction.tex
\section{Introduction}
Image inpainting is the task to fill in the missing part of an image with visually plausible contents. It is one of the most common operations of image editing~\cite{gatys2015texture} and low-level computer visions~\cite{komodakis2006image,hays2007scene}. The goal of image inpainting is to create semantically plausible contents with realistic texture details. The inpainting can be consistent with the original image or different but coherent with the known context. Other than restoring and fixing damaged images, inpainting can also be used to remove unwanted objects, or in the case of guided inpainting, be used to composite with another guide image. In the latter scenario, we often need inpainting to fill in the gaps and remove discontinuities between target region of interest on the guide image and the source context. Image harmonization is also required to adjust the appearance of the guide image such that it is compatible with the source, making the final composition appear natural. 

Traditional image inpainting methods mostly develop texture synthesis techniques to address the problem of hole-filling~\cite{bertalmio2000image,komodakis2006image,wexler2004space,barnes2009patchmatch,bertalmio2003simultaneous,wilczkowiak2005hole}. In~\cite{barnes2009patchmatch}, Barnes et al. proposes the Patch-Match algorithm which efficiently searches the most similar patch to reconstruct the missing regions. Wilczkowiak et al.~\cite{wilczkowiak2005hole} takes further steps and detects desirable search regions to find better match patches. However, these methods only exploit the low-level signal of the known contexts to hallucinate missing regions and fall short of understanding and predicting high-level semantics. Furthermore, it is often inadequate to capture the global structure of images by simply extending texture from surrounding regions. Another line of work in inpainting aims to fill in holes with contents from another guide image by using composition and harmonization~\cite{hays2007scene,tsai2017deep}. The guide is often retrieved from a large database based on similarity with the source image and is then combined with the source. Although these methods are able to propagate high-frequency details from the guide image, they often lead to inconsistent regions and gaps which are easily detectable with human eyes.  

\begin{figure*}[t]
\centering
\setlength\tabcolsep{1pt}
\begin{tabular}{cccccc}
  \includegraphics[width=.16\textwidth]{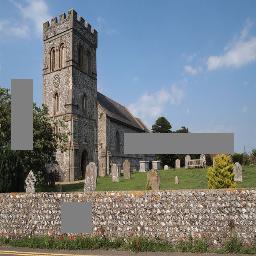}&
  \includegraphics[width=.16\textwidth]{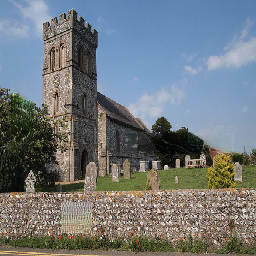}&
  \includegraphics[width=.16\textwidth]{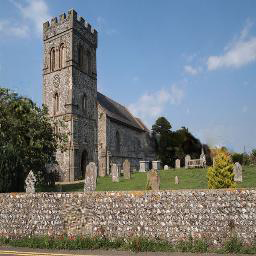} &
    \includegraphics[width=.16\textwidth]{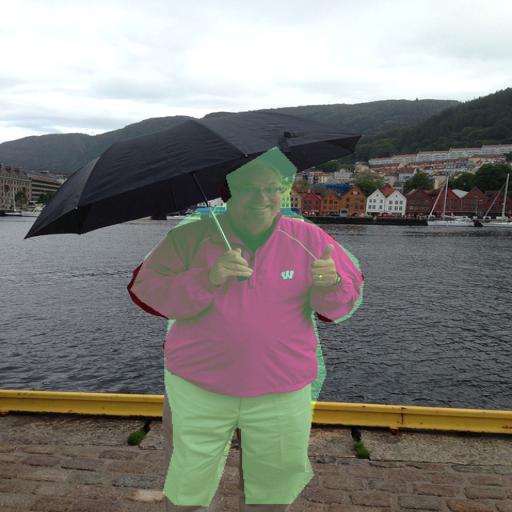}&
  \includegraphics[width=.16\textwidth]{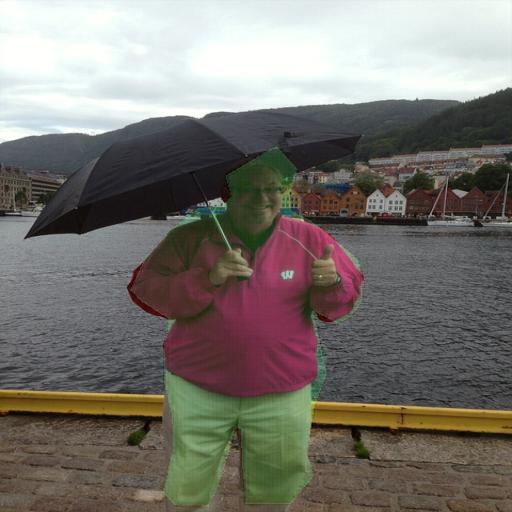}&
  \includegraphics[width=.16\textwidth]{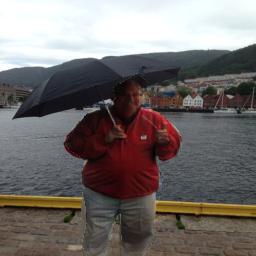} \\
  (a) Input  & (b) GLI~\cite{iizuka2017globally} & (c) Ours & (d) Input  & (e) DH~\cite{tsai2017deep} & (f) Ours  \\
\end{tabular}
\caption{An example of inpainting (left) and harmonzation (right) comparing with state-of-the-art methods. Zoom in for best viewing quality.}
\label{fig:teaser}
\vspace{-15pt}
\end{figure*}

More recently, deep neural networks have shown excellent performance in various image completion tasks, such as texture synthesis and image completion. For inpainting, adversarial training becomes the de facto strategy to generate sharp details and natural looking results~\cite{pathak2016context,yeh2016semantic,li2017generative,yang2017high,iizuka2017globally}. Pathak et al.~\cite{pathak2016context} first proposes to train an encoder-decoder model for inpainting using both reconstruction loss and adversarial loss. In~\cite{yeh2016semantic}, Yeh et al. uses a pre-trained model to find the most similar encoding of a corrupted image and uses the found encoding to synthesize what is missing. In~\cite{yang2017high}, Yang et al. proposes a multi-scale approach and optimizes the hole contents such that neural feature extracted from a pre-trained CNN matches with features of the surrounding context. The optimization scheme improves the inpainting quality and resolution at the cost of computational efficiency. In~\cite{iizuka2017globally}, Iizuka et al. proposes a deep generative model trained with global and local adversarial losses, and can achieve good inpainting performance for mid-size images and holes. However, it requires extensive training (two months as described in the paper), and the results often contain excessive artifacts and noise patterns. Another limitation of ~\cite{yang2017high} and ~\cite{iizuka2017globally} is that they are unable to handle perceptual discontinuity, making it necessary to resort to post-processing (e.g. Poisson blending).

In practice, we foundå that directly training a very deep generative network to synthesize high-frequency details is challenging due to optimization difficulty and unstable adversarial training. As a result, as the network becomes deeper, the inpainting quality may worsen. To overcome this difficulty, we come up with a new training scheme that guides and stabilizes the training of a very deep generative model, which, combining with carefully designed training losses, significantly reduces artifacts and improves the quality of results. The main strategy, referred to as \textbf{B}lock-wise \textbf{P}rocedural \textbf{T}raining (\textbf{BPT}), progressively increases the number of residual blocks and the depth of the network. With BPT, we first train a cGAN-based \textbf{Generator Head} until converging, followed by adding more residual blocks one at a time, which refines and improves the results. This enables us to train a network deeper than~\cite{iizuka2017globally} and generates more realistic-looking details. We also observe that to reduce the noise level, it is essential to steadily reduce the weight of adversarial loss given to the generator. We refer this training scheme as \textbf{A}dversarial \textbf{L}oss \textbf{A}nnealing (\textbf{ALA}). Finally, we also propose two new losses specifically designed for inpainting: the \textbf{P}atch \textbf{P}erceptual \textbf{L}oss (\textbf{PPL}) and \textbf{M}ulti-\textbf{S}cale \textbf{P}atch \textbf{A}dversarial \textbf{L}oss (\textbf{MSPAL}). Experiments show that these losses work better than traditional losses used for inpainting, such as $\ell_2$ and the more general adversarial loss.

To evaluate the proposed approach, we conduct extensive experiments on a variety of datasets. Shown by qualitative and quantitative results, also by the user-study, our approach generates high-quality inpainting results and outperforms other state-of-the-art methods.  We also show that our framework, although being designed for inpainting, can be directly applied on general image translation tasks such as image harmonization and composition, and easily achieve results superior to other methods (Fig.~\ref{fig:teaser}). This enables us to train a multi-task model and use it for interactive guided inpainting, which is a very common and useful image editing scenario. We will describe it in detail in Sec.~\ref{exp:guided}.

In summary, in this paper we present:
\begin{enumerate}
\item A novel and effective framework that generates high-quality results for several image editing tasks like inpainting and harmonization. We also provide a thorough analysis and ablation study about the components in our framework.  
\item Extensive qualitative and quantitative evaluations on a variety of datasets, such as scenes, objects, and faces. We also conduct a user-study to make fair and rigorous comparisons with other state-of-the-art methods. 
\item Results of interactive guided inpainting as a novel and useful application of our approach. 
\end{enumerate}

%% file: 3_Approach.tex
\section{Our Method}

\begin{figure*}[ht!]
\centering
\small
\includegraphics[width=1\textwidth]{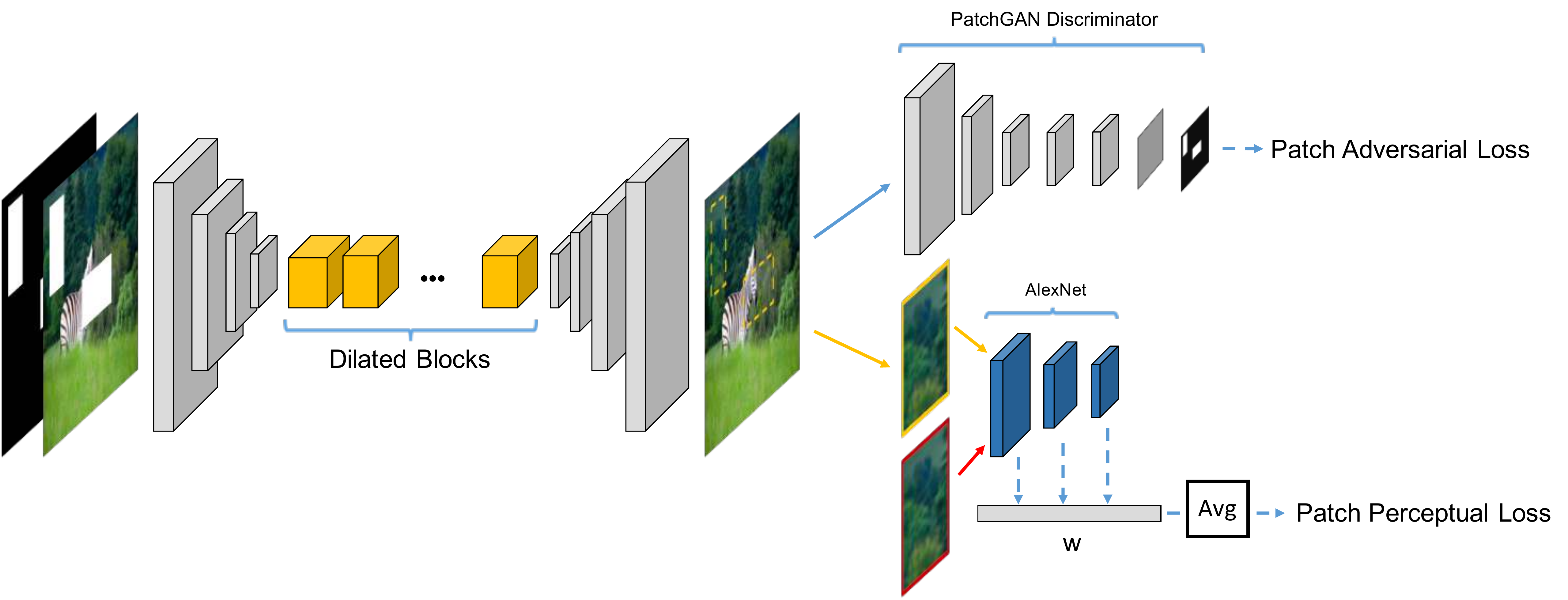}
\caption{Generator head and the training losses. We only illustrate one scale of Patch Adversarial Loss and Patch Perceptual Loss. Note that to compute the Patch Adversarial Loss, we need to use the mask to find out which patch overlaps with the hole.}
\label{fig:model}
\vspace{-15pt}
\end{figure*}

Our method consists of training a generator head as initialization and optimizing additional residual blocks as refinement. The generator head is trained for inpainting and is based on conditional GAN framework. Vanilla cGAN for image inpainting consists of a Generator $G$ and a Discriminator $D$. The generator learns to predict the hole contents and restore the complete image, while the discriminator learns to distinguish real images from generated images. Using the original image as ground truth, the model is trained in a self-supervised manner via the following minimax objective:
\begin{eqnarray}
\min\limits_G \max\limits_D E_{(s,x)}[\log D(s,x)] + E_s[\log (1-D(s,G(s)))].
\end{eqnarray}
Here $s$ and $x$ are the incomplete image as input and the original image as target. $G(s)$ is the output of the generator which could be the hole content or the complete image. In the case when $G(s)$ restores the complete image, only the contents inside the hole are kept to combine with the known regions of $s$.

\subsection{The Generator Head}
\label{sec:resnet_head}

Existing works have investigated different architectures of $G$, most notably the encoder-decoder style of~\cite{pathak2016context} and the FCN style of~\cite{iizuka2017globally}. ~\cite{iizuka2017globally} shows that FCN generates higher-quality and less blurred results comparing with encoder-decoder, largely because the network used is much deeper and is fully convolutional. In contrast, the encoder-decoder in~\cite{pathak2016context} uses an intermediate bottleneck, fully-connected layer, which results in size reduction and information loss. 

Similar to~\cite{iizuka2017globally}, our generator head is based on FCN and leverage the many properties of convolutional neural networks, such as translation invariance and parameter sharing. However, a major limitation of FCN is the constraint of the receptive field size. Given the convolution layers are locally connected, pixels far away from the hole carry no influence on the hole. We propose several modifications to alleviate this drawback. First, we build our network with three components: a down-sampling front end to reduce the size, followed by multiple residual blocks~\cite{he2016deep}, and an up-sampling back end to restore the full dimension. Using down-sampling increases the receptive field of the residual blocks and makes it easier to learn the transformation at a smaller size. Second, we stack multiple residual blocks to further enlarge the view at later layers. Finally, we adopt the dilated convolution~\cite{yu2015multi} in all residual blocks, with the dilation factor set to 2. Dilated convolutions use spaced kernels, making it compute each output value with a wider view of input without increasing the number of parameters and computational burden. From experiments, we observe that increasing the size of the receptive field, especially by using dilated convolution, is critical for enhanced inpainting quality. In contrast, other image translation tasks such as super-resolution, denoising, etc., usually rely more on local statistics rather than global context.

The detail of our architecture is as follows: the down-sampling front end consists of three convolutional layers, each with a stride of 2. The intermediate part contains 9 residual blocks, and the up-sampling back-end consists of three transposed convolution, also with a stride of 2. Each convolutional layer is followed by batch normalization (BN) and ReLu activation, except for the last layer which outputs the image. Similar architecture without dilated convolution has been used in~\cite{wang2017high} for image synthesis. Comparing with~\cite{iizuka2017globally}, our receptive field is much larger, as we adopted more down-sampling and dilated layers. This makes us generate better results with fewer artifacts. As ablation study, we compare different types of layers and architectures in Sec.~\ref{exp:study}.

\subsection{The Training Losses}
Different losses have been used to train an inpainting network. These losses can be cast into two categories. The first category which can be referred to as \textit{similarity loss}, is used to measure the similarity between the output and the original image. The second category which we refer to as the \textit{realism loss}, is used to measure how realistic-looking is the output image. We summarize the losses used in different inpainting methods in Table~\ref{table:losses}. 

\begin{table}[h!]
\begin{center}
\resizebox{.5\textwidth}{!}{%
{\tiny
  \begin{tabular}{ l  c  c }
    \hline
    \textbf{Method} & \textbf{Similarity Loss} &  \textbf{Realism Loss} \\ \hline
    \emph{CE~\cite{pathak2016context}} & $\ell_2$ & Global Adversarial Loss\\ \hline
    \emph{GLI~\cite{iizuka2017globally}} & $\ell_2$ & Global and Local Adversarial Loss \\ \hline
    \emph{Ours} & Patch Perceptual Loss & Multi-Scale Patch Adversarial Loss \\ \hline
    \hline
  \end{tabular}}
  }
  \end{center}
  \caption{Comparison of training losses used in different methods.}
  \label{table:losses}
\end{table}

\noindent\textbf{Patch Perceptual Loss} As shown in Table~\ref{table:losses}, using $\ell_2$ as reconstruction loss to measure the difference between the output and the original image has been the default choice of previous inpainting methods. However, it is known that $\ell_2$ loss does not correspond well to human perception of visual similarity (Zhang et al.~\cite{zhang2018unreasonable}). This is because using $\ell_2$ assumes each output pixel is conditionally independent of all others, which is not the case. An example is that blurring an image leads to small changes in terms of $\ell_2$ distance but causes significant perceptual difference. Recent research suggests that a better metric for perceptual similarity is the internal activations of deep convolutional networks, usually trained on a high-level image classification task. Such loss is called ``perceptual loss'', and is used in various tasks such as neural style transfer~\cite{gatys2016image}, image super-resolution~\cite{johnson2016perceptual}, and conditional image synthesis~\cite{dosovitskiy2016generating,chen2017photographic}.

Based on this observation, we propose a new ``patch perceptual loss'' as the substitute of the $\ell_2$ losses. Traditional perceptual loss typically uses VGG-Net and computes the $\ell_2$ distance of the activations on a few feature layers. Recently,~\cite{zhang2018unreasonable} trained a specific perceptual network based on AlexNet to measure the perceptual differences between two image patches, making it an ideal candidate for our task. The perceptual network computes the activations and sums up the $\ell_2$ distances across all feature layers, each scaled by a learned weight. Furthermore, taking both the local view and the global view into account, we compute PPL at two scales. Local PPL considers the local hole patch, while the global PPL slightly zooms out to cover a larger contextual area. More formally, our PPL is defined as:
\begin{eqnarray}
&& \sum\limits_{p=p_1,p_2}PPL(G(s)_p, x_p) \\ \nonumber
&&= \sum\limits_{p=p_1,p_2}\sum\limits_l\frac{1}{H_lW_l}\sum\limits_{h,w}\parallel w_l^T\odot(\hat{F}(x_p)^l_{hw}-\hat{F}(G(s)_p)^l_{hw})\parallel^2_2.
\end{eqnarray}

Here $p$ refers to the hole patch. $\hat{F}$ is the AlexNet and $l$ is the feature layer. $w_l$ is the layer-wise learned weight. Ablation study in Sec.~\ref{exp:study} shows that our patch perceptual loss gives better inpainting quality than traditional $\ell_2$. 

\noindent\textbf{Multi-Scale Patch Adversarial Loss} Adversarial losses are given by trained discriminators to discern whether an image is real or fake. The global adversarial loss of~\cite{pathak2016context} takes the entire image as input and outputs a single real/fake prediction which does not consider local realism of holes. The additional local adversarial loss of~\cite{iizuka2017globally} adds another discriminator specifically for the hole, but it requires the hole to be fixed in shape and size during training to fit the local discriminator. To consider both the global view and the local view, and to be able to use multiple holes of arbitrary shape, we propose to use PatchGAN discriminators~\cite{isola2016image} at three scales of image resolutions. The discriminator at each scale is identical and only the input is a differently scaled version of the \textit{entire image}. Each discriminator is a fully convolutional PatchGAN and outputs a vector of real/fake predictions and each value corresponds to a local image patch. In this way, the discriminators are trained to classify global and local patches across the image at multiple scales, and it enables us to use multiple holes of arbitrary shapes since now the input is the entire image rather than the hole itself. However, since the output image is the composition of synthesized holes and known background, directly using PatchGAN is problematic because it does not differentiate between the hole patches (\textit{fake}) and the background patches (\textit{real}). To address this issue, when computing the PatchGAN loss, only the patches that overlap with the holes are labeled as fake. More formally, our Patch-wise Adversarial Loss is defined as: 
\begin{eqnarray}
&&\min\limits_G\max\limits_{D_1, D_2, D_3}\sum\limits_{k=1,2,3} L_{GAN}(G,D_k)  \\ \nonumber
&&=\sum\limits_{k=1,2,3}E_{(s_k,x_k)}[\log (s_k,x_k)] + E_{s_k}[\log (Q_k-D_k(s_k,G(s_k)))].
\end{eqnarray}
\label{eqn:adversarial_loss}
Here $k$ refers to the image scale, and $Q_k$ is a patch-wise real/fake vector, based on whether the patch overlaps with the holes. Using multiple GAN discriminators at the same or different image scale has been proposed in unconditional GANs~\cite{durugkar2016generative} and conditional GANs~\cite{wang2017high}. Here we extend the design to take into account the holes, which is critical for obtaining semantically and locally coherent image completion results.

To summarize, our full objective combines both losses is therefore defined as:

\begin{eqnarray}
\min\limits_G((\max\limits_{D_1, D_2, D_3}\lambda_{Adv}\sum\limits_{k=1,2,3} L_{GAN}(G,D_k)) \\ \nonumber
+\lambda_{PPL}\sum\limits_{k=1,2}PPL_k(G(s)_p, x_p)).
\end{eqnarray}

We use $\lambda_{Adv}$ and $\lambda_{PPL}$ to control the importance of the two terms. In our experiment, we set initial $\lambda_{Adv}=1$ and $\lambda_{PPL}=10$ as used in~\cite{pathak2016context,wang2017high}. As training progresses, we gradually decrease the weight of $\lambda_{Adv}$ based on adversarial loss annealing (Sec.~\ref{sec:procedural}).

Fig.~\ref{fig:model} illustrates the architecture of our generator head and the training losses.

\subsection{Blockwise Procedural Training with Adversarial Loss Annealing}
\label{sec:procedural}
Our experiments show that using the described generator head and training losses already generate results of decent quality. An intuitive effort to improve the results would be to stack more intermediate residual blocks to further expand the receptive view and increase the expressiveness of the model. However, we found that directly stacking more residual blocks makes it more difficult to stabilize the training. As the parameter space becomes much larger, it is also more challenging to find local optimum. In the end, the inpainting quality deteriorates as the model depth increases.

\begin{figure}[t]
\centering
\small
\includegraphics[width=.45\textwidth]{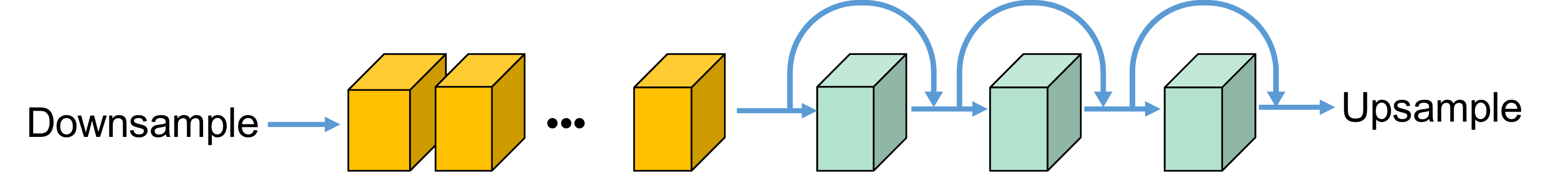}
\caption{Illustration of block-wise procedural training. The yellow residual blocks refer to the generator head which is trained first. The green residual blocks are progressively added one at a time. We also draw the skip connections between the already trained residual blocks and the up-sampling back end.}
\label{fig:procedural}

\end{figure}

To address the challenge of training a deeper network, we propose to use procedural block-wise training to gradually increase the depth of the inpainting network. More specifically, we begin by training the generator head until it converges. Then, we add a new residual block after the already trained residual blocks, right before the back-end up-sampling layers. In order to smoothly introduce the new residual block without breaking the trained model, we add another skip path from the already trained residual blocks to the up-sampling layers. Initially, the weight of the skip path is set to 1, while the weight of the path containing the new block is set to 0. This essentially makes the initial network identical to the already trained network. We then slowly decrease the weight of the skip path and increase the weight of the new residual block as training progresses. In this way, the newly introduced residual block is trained to be a fine-tuning component, which adds another layer of fine details to the original results. This step are repeated multiple times, and each time we deepen the network by adding a new residual block. In our experiments, we found that the results improve significantly over generator head, after training with the first residual block added. The output becomes stabilized after three residual blocks, and very little changes can be detected if more residual blocks are brought in. We illustrate the procedural training process in Fig.~\ref{fig:procedural}.

The block-wise procedural training has several benefits. First, it guides the training process of a very deep generator. Starting with the generator head and gradually fine-tuning with more residual blocks makes it easier to discover the mapping between the incomplete image and the complete image, even though the search space is huge given the diversity of natural images and the randomness of holes. Another benefit is training efficiency, as we found decoupling the training of the generator head and the fine-tuning of additional residual blocks requires significantly less training time comparing with training the network all at once. Similar idea of progressive training has been used in Progressive GAN~\cite{karras2017progressive}. However, the task and the model are different in our case.

\noindent\textbf{Adversarial Loss Annealing} During training, the generator adversarial loss updates the generator weight if the discriminator successfully detects the generated image as fake:
\begin{eqnarray}
\sum\limits_{k=1,2,3}E_s[\log (\bar{Q}-D_k(s_k,G(s_k)))].
\end{eqnarray}
Note that here $\bar{Q}$ reverses $Q$ of~\ref{eqn:adversarial_loss} as this is the loss to the generator. We observe that the generator adversarial loss becomes dominant over PPL as training progresses, as the discriminator becomes increasingly good at detecting fake images during training. This is less of a problem for image synthesis tasks. However, for the inpainting task which requires the output to be faithful to the original image, the outcome is that the generator deliberately adds noise patterns to confuse the discriminators, leading to artifacts or incorrect textures. Based on this observation, we propose to use adversarial loss annealing, which decreases the weight of the adversarial loss of the generator at the time of adding new residual blocks. More formally, let the initial weight of the generator adversarial loss be $\lambda^0_{adv}$, and the weight of the generator adversarial loss be $\lambda^i_{adv}$ after adding the $i_{th}$ residual block. We found that simply decay the weight linearly by setting $\lambda^i_{G_{adv}}=10^{-i}\lambda^0_{G_{adv}}$ gives results significantly better than using constant weight. In Sec.~\ref{sec:results}, we analyze the effect of ALA in more detail.

Finally, we summarize the discussed training schemes in Alg.~\ref{algo}.
\begin{algorithm}
\caption{Training the Inpainting Network}\label{algo}
\begin{algorithmic}[1]
\State Set batch size to 8 and basic learning rate to $lr_0\gets 0.0002$.
\State Set $\lambda_{PPL}\gets 10$ and $\lambda^0_{{adv}}\gets 1$.
\State Train the Generator Head $G_0$ using MSPAL and PPL (\ref{eqn:adversarial_loss}) for 150,000 iterations.
\For{i=1 to 3}
\State Add the skip path and the residual block $r_i$.
\State Set $lr_i\gets 10^{-i} lr_0$ and $\lambda^i_{G_{adv}}\gets 10^{-i}\lambda^0_{G_{adv}}$.
\State Train the generator $G_3$ with the added $r_i$ for 1,500 iterations.
\EndFor
\State \textbf{return} $G_3$ 
\end{algorithmic}
\end{algorithm}

%% file: 4_Experiments.tex
\section{Results}
\label{sec:results}
In this section, we first describe our datasets and experiment setting (Sec.~\ref{exp:setup}). We then provide quantitative and qualitative comparisons with several methods, and also report a subjective human perceptual test based on user-study (Sec.~\ref{exp:comparison}). In Sec.~\ref{exp:study}, we conduct several ablation study about the design choice of the framework. Finally, we show how our method can be applied to image harmonization and interactive guided inpainting (Sec.~\ref{exp:guided}).

\subsection{Experiment Setup}
\label{exp:setup}
We evaluate our inpainting method on ImageNet~\cite{russakovsky2015imagenet}, COCO~\cite{lin2014microsoft}, Place2~\cite{zhou2016places} and CelebA~\cite{liu2015faceattributes}. CelebA consists of 202,599 face images with various viewpoints and expressions. For a fair comparison, we follow the standard split with 162,770 images for training, 19,867 for validation and 19,962 for testing.

In order to compare with existing methods, we train on images of size 256x256. We also train another network at a larger scale of 512x512 to demonstrate its ability to handle higher resolutions. For each image, we apply a mask with a single hole or multiple holes placed at random locations. The size of hole is between 1/4 and 1/2 of the image's size. We set the batch size to 8, and regardless of the actual dataset size, we train 150,000 iterations for the generator head and another 1,500 iterations for each additional residual block. For each dataset, the training takes around 2 days to finish on a single Titan X GPU, which is significantly less time than~\cite{iizuka2017globally}.

\begin{figure*}[!ht]
\centering
\setlength\tabcolsep{1pt}
\begin{tabular}{ccccccc}
\includegraphics[width=.14\textwidth]{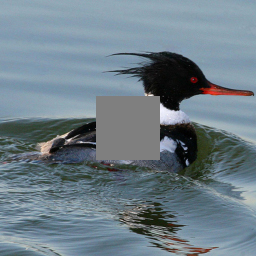}&
\includegraphics[width=.14\textwidth]{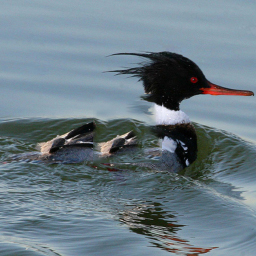}&
\includegraphics[width=.14\textwidth]{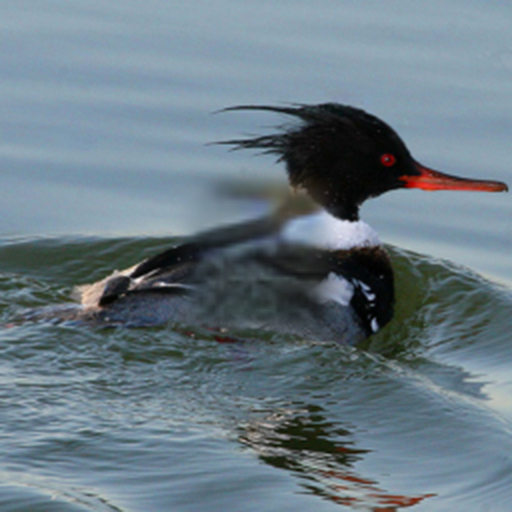}&
\includegraphics[width=.14\textwidth]{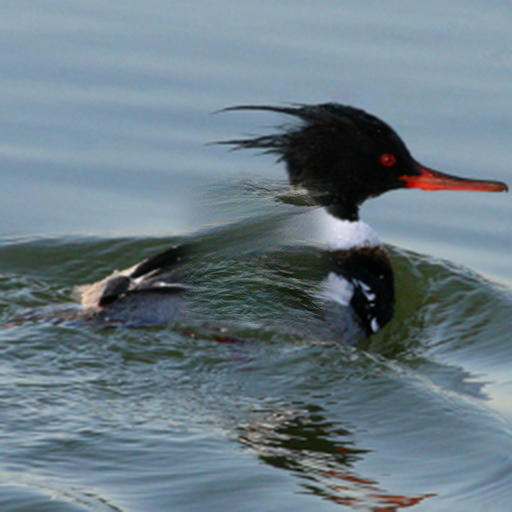}&
\includegraphics[width=.14\textwidth]{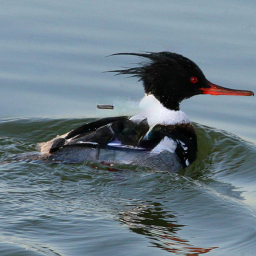}&
\includegraphics[width=.14\textwidth]{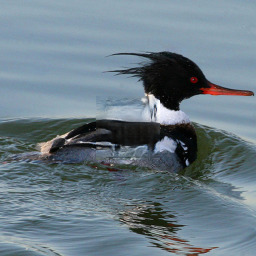}&
\includegraphics[width=.14\textwidth]{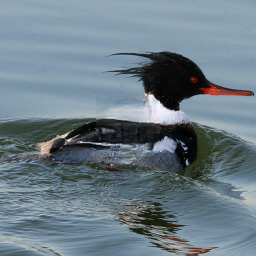}\\
\includegraphics[width=.14\textwidth]{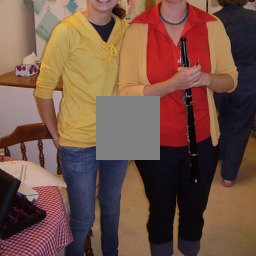}&
\includegraphics[width=.14\textwidth]{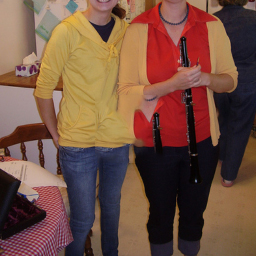}&
\includegraphics[width=.14\textwidth]{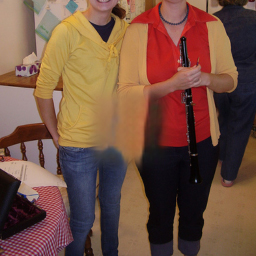}&
\includegraphics[width=.14\textwidth]{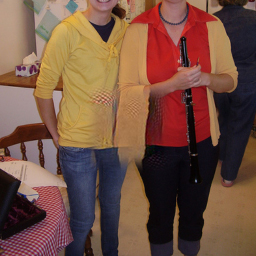}&
\includegraphics[width=.14\textwidth]{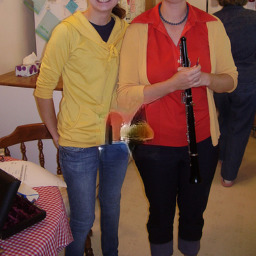}&
\includegraphics[width=.14\textwidth]{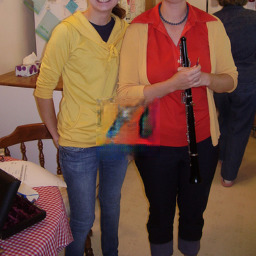}&
\includegraphics[width=.14\textwidth]{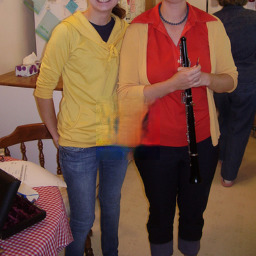}\\
\includegraphics[width=.14\textwidth]{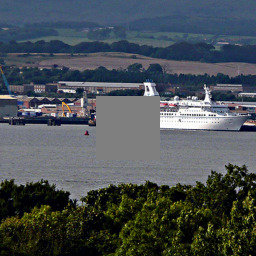}&
\includegraphics[width=.14\textwidth]{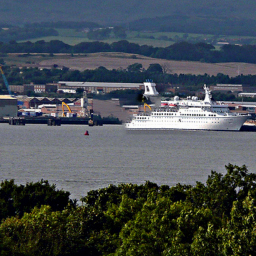}&
\includegraphics[width=.14\textwidth]{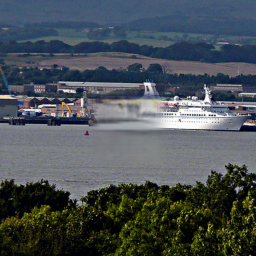}&
\includegraphics[width=.14\textwidth]{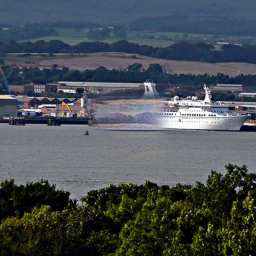}&
\includegraphics[width=.14\textwidth]{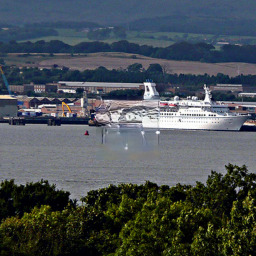}&
\includegraphics[width=.14\textwidth]{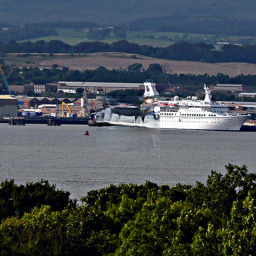}&
\includegraphics[width=.14\textwidth]{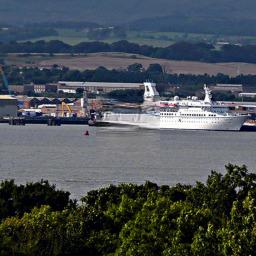}\\
\includegraphics[width=.14\textwidth]{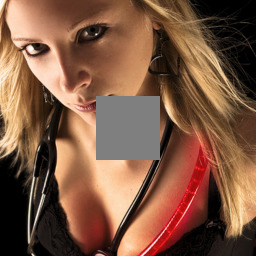}&
\includegraphics[width=.14\textwidth]{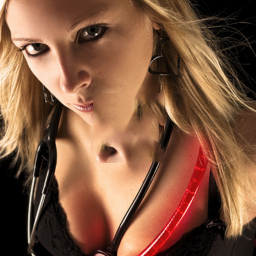}&
\includegraphics[width=.14\textwidth]{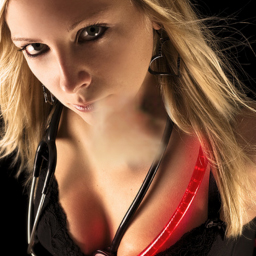}&
\includegraphics[width=.14\textwidth]{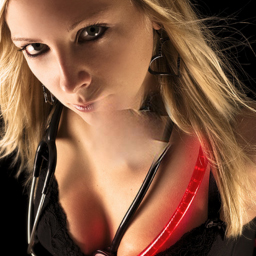}&
\includegraphics[width=.14\textwidth]{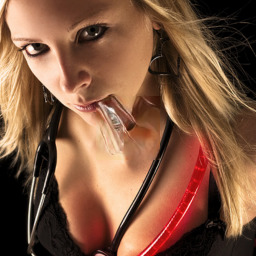}&
\includegraphics[width=.14\textwidth]{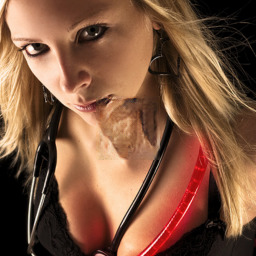}&
\includegraphics[width=.14\textwidth]{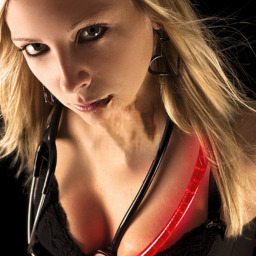}\\
(a) Input & (b) CAF~\cite{barnes2009patchmatch} & (c) CE~\cite{pathak2016context} & (d) NPS~\cite{yang2017high} & (e) GLI~\cite{iizuka2017globally} & (f) GH & (g) Final \\
\end{tabular}
\caption{Fixed hole result comparisons. GH are results generated by generator head, and Final are results generated with block procedural training. All images have original size 256x256. Zoom in for best viewing quality.}
\label{fig:fixed}
\end{figure*}

\begin{figure*}[!ht]
\centering
\small
\setlength\tabcolsep{1pt}
\begin{tabular}{ccccc}
\includegraphics[width=.2\textwidth]{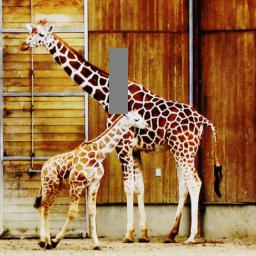}&
\includegraphics[width=.2\textwidth]{figures/random/000000153299_pm.jpg}&
\includegraphics[width=.2\textwidth]{figures/random/000000153299_sig.jpg}&
\includegraphics[width=.2\textwidth]{figures/random/000000153299_g.jpg}&
\includegraphics[width=.2\textwidth]{figures/random/000000153299.jpg}\\
\includegraphics[width=.2\textwidth]{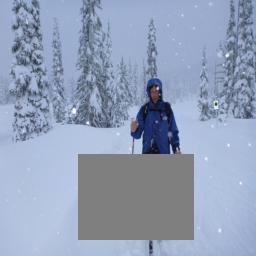}&
\includegraphics[width=.2\textwidth]{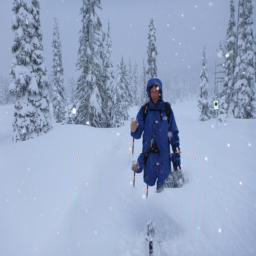}&
\includegraphics[width=.2\textwidth]{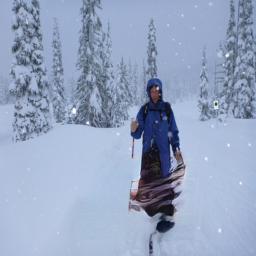}&
\includegraphics[width=.2\textwidth]{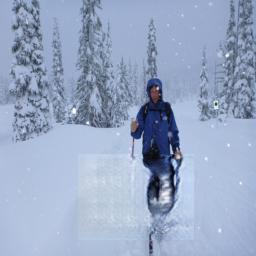}&
\includegraphics[width=.2\textwidth]{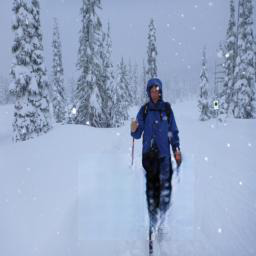}\\
\includegraphics[width=.2\textwidth]{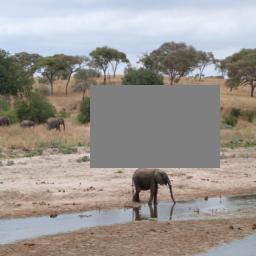}&
\includegraphics[width=.2\textwidth]{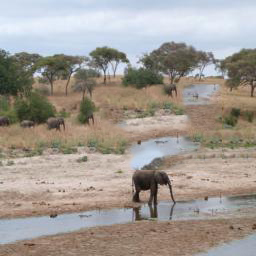}&
\includegraphics[width=.2\textwidth]{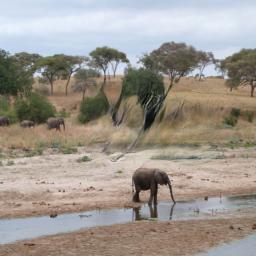}&
\includegraphics[width=.2\textwidth]{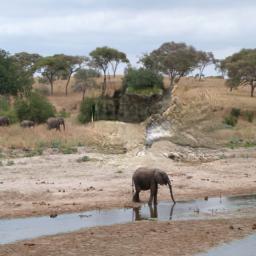}&
\includegraphics[width=.2\textwidth]{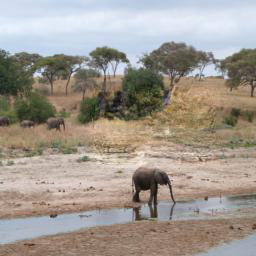}\\
(a) Input & (b) CAF~\cite{barnes2009patchmatch} & (c) GLI~\cite{iizuka2017globally} & (d) GH & (e) Final \\
\end{tabular}
\caption{Random hole completion comparison. GH are results generated by generator head, and Final are results generated with block procedural training. All images have original size 256x256. Zoom in for best viewing quality.}
\label{fig:random}
\end{figure*}

\subsection{Comparison with Existing Methods}
\label{exp:comparison}
For ImageNet, COCO and Place2, we compare our results with Content-Aware Fill (CAF)~\cite{barnes2009patchmatch}, Context Encoder (CE)~\cite{pathak2016context}, Neural Patch Synthesis (NPS)~\cite{yang2017high} and Global Local Inpainting (GLI)~\cite{iizuka2017globally}. For CE, NPS, and GLI, we use pre-trained models made available by the authors. CE and NPS are only trained with holes with fixed size and location (image center), while CAF and GLI can handle arbitrary holes. For fair comparisons, we evaluate on both settings. For center hole completion, we compare with CAF, CE, NPS and GLI on ImageNet~\cite{russakovsky2015imagenet} test images. For random hole completion, we compare with CAF and GLI using test images from COCO and Place2. Only GLI applies Poisson Blending as post-processing. We also show the results by our \textit{Generator Head} and compare. The images shown in Fig.~\ref{fig:fixed},~\ref{fig:random} are randomly sampled from the test set. 

We see that, while CAF can generate realist-looking details using patch propagation, they often fail to capture the global structure or synthesize new contents. CE's result is significantly blurrier as it can only inpaint low-resolution images. NPS's results heavily depend on CE's initialization. Our approach is much faster and generates results of higher visual quality most of the time. Comparing with GLI, our results do not need post-processing, and are more coherent with fewer artifacts. Our final results also show significant improvement over Generator Head, demonstrating the effectiveness block-wise procedural fine-tuning.

\begin{figure*}[!ht]
\centering
\small
\setlength\tabcolsep{1pt}
\begin{tabular}{cccccc}
\includegraphics[width=.16\textwidth]{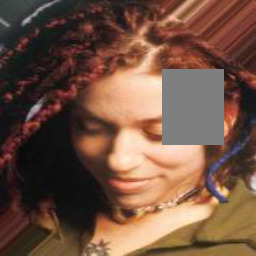}&
\includegraphics[width=.16\textwidth]{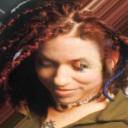}&
\includegraphics[width=.16\textwidth]{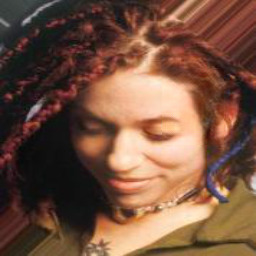}&
\includegraphics[width=.16\textwidth]{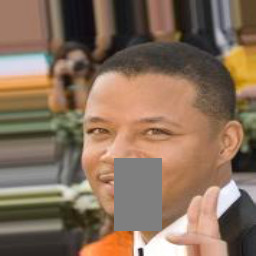}&
\includegraphics[width=.16\textwidth]{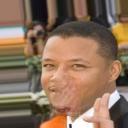}&
\includegraphics[width=.16\textwidth]{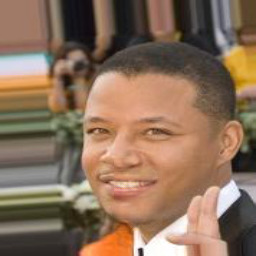}\\
\includegraphics[width=.16\textwidth]{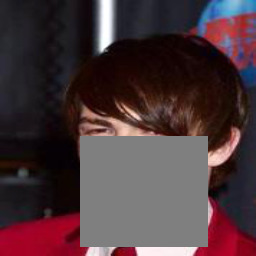}&
\includegraphics[width=.16\textwidth]{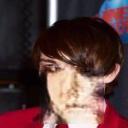}&
\includegraphics[width=.16\textwidth]{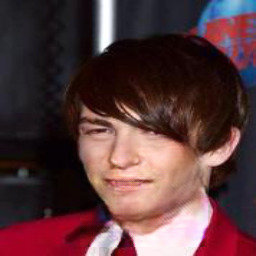}&
\includegraphics[width=.16\textwidth]{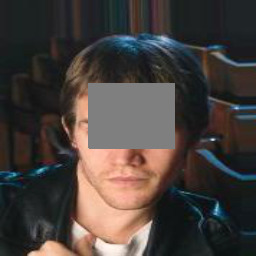}&
\includegraphics[width=.16\textwidth]{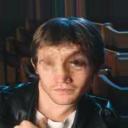}&
\includegraphics[width=.16\textwidth]{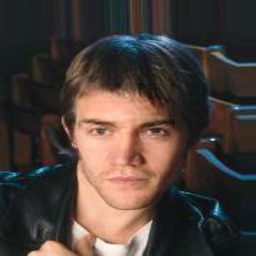}\\
(a) Input & (b) GFC~\cite{li2017generative} & (c) Ours & (d) Input & (e) GFC~\cite{li2017generative} & (f) Ours \\
\end{tabular}
\caption{Face completion results comparing with~\cite{li2017generative}.}
\label{fig:face}
\vspace{-10pt}
\end{figure*}

For CelebA, we compare our results with Generative Face Completion~\cite{li2017generative} in Fig.~\ref{fig:face}. Since~\cite{li2017generative}'s model inpaints images of 128x128, we directly up-sample the results to 256x256 to compare. The images shown are also chosen at random. We can see that while~\cite{li2017generative} is an approach specific for face inpainting, our method for general inpainting tasks actually generates much better face completion results.

\noindent\textbf{Quantitative Evaluation} Table~\ref{table:numerical} shows quantitative comparison between CAF, CE, GLI and our approach. The values are computed based on a random subset of 200 images selected from the test set. Both $\ell_1$ and $\ell_2$ errors are computed with pixel values normalized between $[0,1]$ and are summed up using all pixels of the image. We can see that our method performs better than other methods in terms of SSIM and $\ell_1$ error. For $\ell_2$, CAF and GLI have smaller errors. This is understandable as our model is trained with perceptual loss rather than $\ell_2$ loss. Besides, $\ell_2$ loss awards averaging color values and does not faithfully reflect perceptual quality. From the numerical values, we can also see that procedural fine-tuning reduces the errors from the generator head. 

\begin{table}[!ht]
\begin{center}
\resizebox{.5\textwidth}{!}{%
{\tiny
  \begin{tabular}{ l  c c c }
    \hline
    \textbf{Method} & \textbf{Mean $\ell_1$ Error} & \textbf{Mean $\ell_2$ Error} & \textbf{SSIM} \\ \hline
    \multirow{2}{*}{CAF~\cite{barnes2009patchmatch}} & 968.8 & \textbf{209.5} & 0.9415\\ \cline{2-4}
    & 1660 & \textbf{363.5} & 0.9010\\ \hline
    \multirow{2}{*}{CE~\cite{pathak2016context}} & 2693 & 545.8 & 0.7719\\ \cline{2-4}
    & N/A & N/A & N/A\\ \hline
    \multirow{2}{*}{GLI~\cite{iizuka2017globally}} & 868.6 & 269.7 & 0.9452 \\ \cline{2-4}
    & 1640 & 378.7 & 0.9020\\ \hline
    \multirow{2}{*}{Ours (Generator Head)} & 913.8 & 245.6 & 0.9458  \\ \cline{2-4}
    & 1629 & 439.4 & 0.9073\\ \hline
    \multirow{2}{*}{Ours (Final)} & \textbf{838.3} & 253.3 & \textbf{0.9486} \\ \cline{2-4}
    & \textbf{1609} & 427.0 & \textbf{0.9090}\\ \hline\hline

  \end{tabular}}
  }
  \end{center}
  \caption{Numerical comparison between CAF, CE and GLI, our generator head results and our final results. Up/down are results of center/random region completion. Note that for SSIM, larger values mean greater similarity in terms of content structure and indicate better performance.}
  \label{table:numerical}
\end{table}

\noindent\textbf{User Study}
To more rigorously evaluate the performance, we conduct a user study based on the random hole results. We asked for feedback from 20 users, by giving each user 30 tuples of images to rank. Each tuple contains results of NPS, GLI, and ours. The user study shows that our method achieves highest scores, outperforming NPS and GLI by a large margin. Among 600 total comparisons, our results are ranked the best 72.3\% of the time. Our results are overwhelmingly better than NPS, as our results are ranked better 95.8\% of the time. Comparing with GLI, our results are ranked better 73.5\% of the time and are ranked the same 15.5\% of the time. 

\subsection{Ablation Study}
\label{exp:study}

\begin{figure*}[!ht]
\centering
\small
\setlength\tabcolsep{1pt}
\begin{tabular}{ccccc}
\includegraphics[width=.2\textwidth]{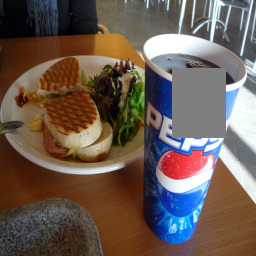}&
\includegraphics[width=.2\textwidth]{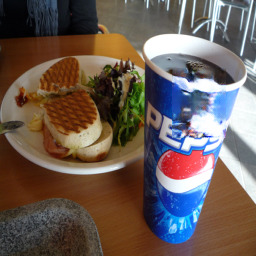}&
\includegraphics[width=.2\textwidth]{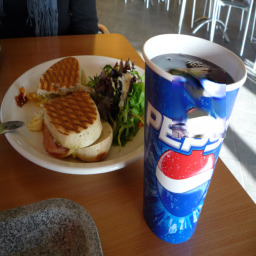}&
\includegraphics[width=.2\textwidth]{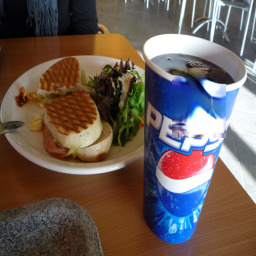}&
\includegraphics[width=.2\textwidth]{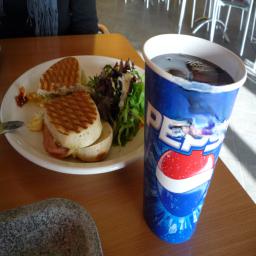}\\
(a) & (b) & (c) & (d) & (e)  \\
\end{tabular}
\caption{Visual comparison of a result using different types of convolutional layers. (a) Input; (b) Original Conv; (c) Interpolated Conv; (d) Dilated+Interpolated Conv; (e) Dilated Conv (ours).}
\label{fig:conv}
\end{figure*}

\noindent\textbf{Comparison of Convolutional Layer} Choosing the proper convolutional layer improves the inpainting quality and also reduces the noise. We consider three types of convolutional layers: original, dilated~\cite{yu2015multi} and interpolated~\cite{odena2016deconvolution}. We train three networks under the same setting to specifically test their effects. We observed that using dilation significantly improves the inpainting quality while using interpolated convolution tends to generate over-smoothed results. Fig.~\ref{fig:conv} shows a qualitative comparison. 

\begin{figure*}[!ht]
\centering
\small
\setlength\tabcolsep{1pt}
\begin{tabular}{cccccc}
\includegraphics[width=.17\textwidth]{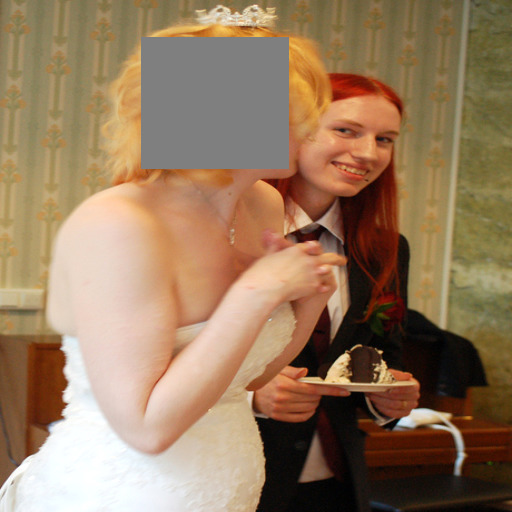}&
\includegraphics[width=.17\textwidth]{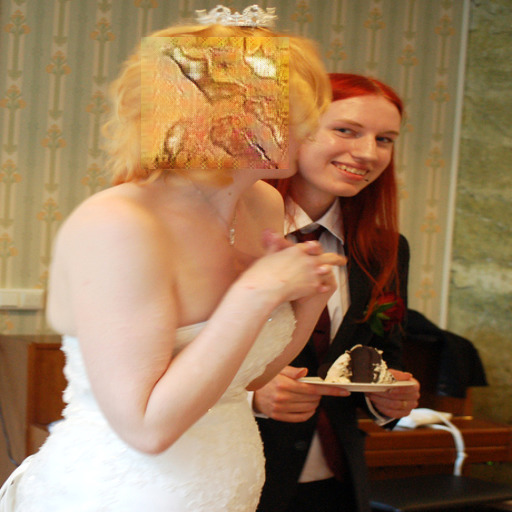}&
\includegraphics[width=.17\textwidth]{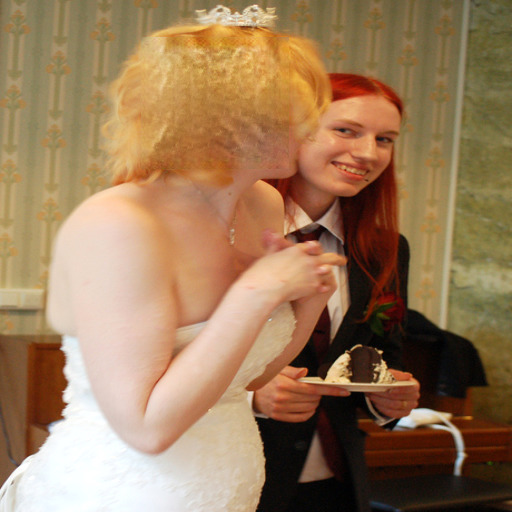}&
\includegraphics[width=.17\textwidth]{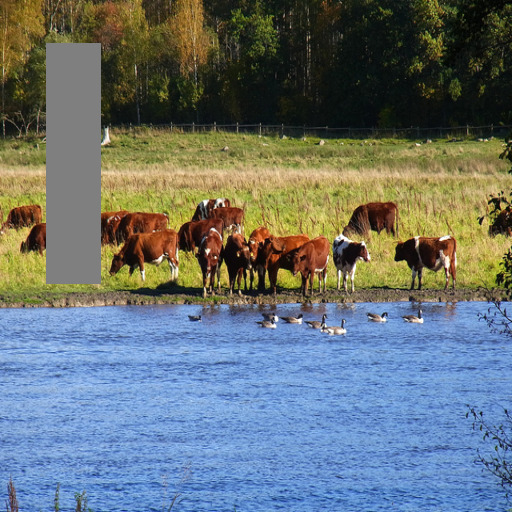}&
\includegraphics[width=.17\textwidth]{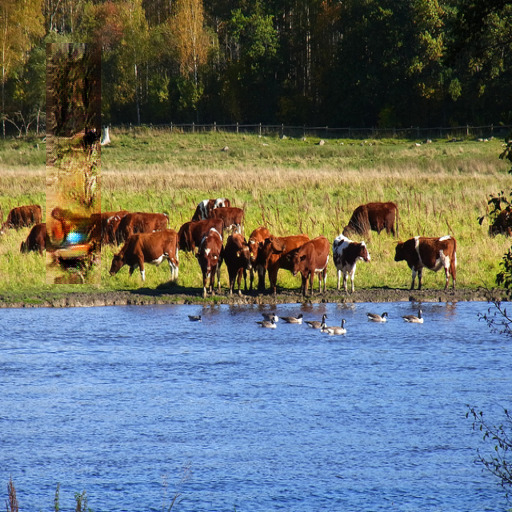}&
\includegraphics[width=.17\textwidth]{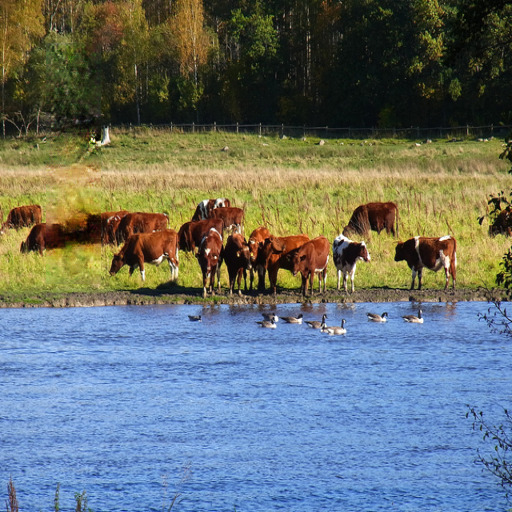}\\
(a) Input & (b)  & (c)  & (d) Input & (e)  & (f)  \\
\end{tabular}
\caption{Visual comparison of results by directly training a network of 12 blocks ((b),(e)) and procedural training ((c), (f)).}
\label{fig:proc}
\end{figure*}

\noindent\textbf{Effect of Procedural Training} Qualitative comparisons in Sec.~\ref{exp:comparison} shows that procedurally adding residual blocks to fine-tune improves the results. However, one may wonder whether we can directly train a deeper network. To find out, we trained a network with 12 residual blocks from scratch, keeping all other hyper-parameters the same. Fig.~\ref{fig:proc} shows two examples of the result comparisons. We found that further increasing the number of residual blocks has a negative effect on inpainting, possibly due to a variety of factors such as gradient vanishing, optimization difficulty, etc. This demonstrates the necessity and importance of the procedural training scheme.

\begin{figure*}[!ht]
\centering
\small
\setlength\tabcolsep{1pt}
\begin{tabular}{cccccc}
\includegraphics[width=.17\textwidth]{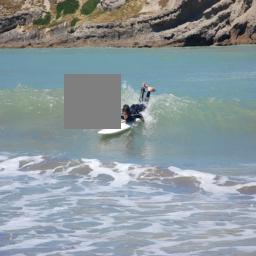}&
\includegraphics[width=.17\textwidth]{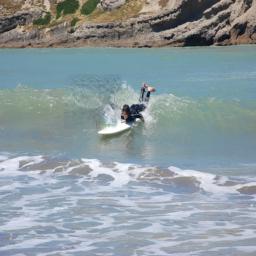}&
\includegraphics[width=.17\textwidth]{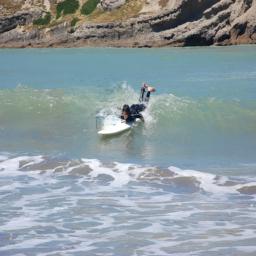}&
\includegraphics[width=.17\textwidth]{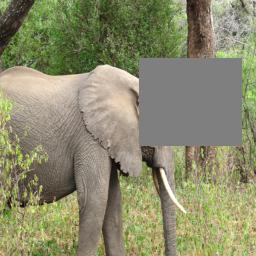}&
\includegraphics[width=.17\textwidth]{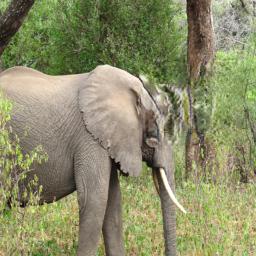}&
\includegraphics[width=.17\textwidth]{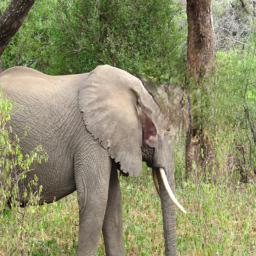}\\
(a) Input & (b) w/o ALA & (c) ALA & (d) Input & (e) w/o ALA & (f) ALA \\
\end{tabular}
\caption{Visual comparison of results by training training without and with ALA.}
\label{fig:ala}
\end{figure*}

\noindent\textbf{Effect of Adversarial Loss Annealing} We also investigate the effect of adversarial loss annealing and train another network without using ALA. Visually, we observe that using adversarial loss annealing reduces the noise level without sacrificing the overall sharpness and realism. We show two randomly selected examples in Fig.~\ref{fig:ala} to illustrate the effects. 

\begin{figure*}[h!]  
\centering  
\small  
\setlength\tabcolsep{1pt}
\begin{tabular}{cccccc} 
\includegraphics[width=.17\textwidth]{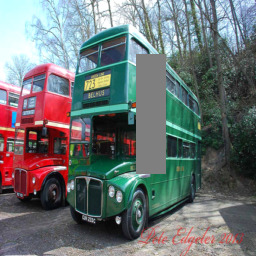}& 
\includegraphics[width=.17\textwidth]{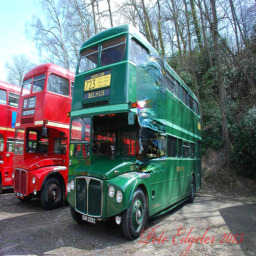}& 
\includegraphics[width=.17\textwidth]{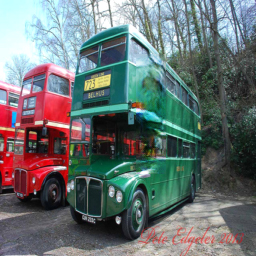}& 
\includegraphics[width=.17\textwidth]{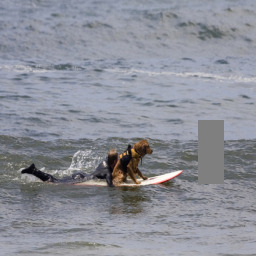}& 
\includegraphics[width=.17\textwidth]{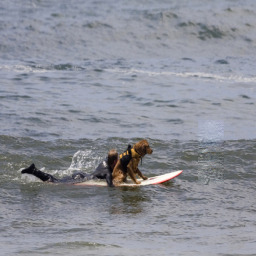}& 
\includegraphics[width=.17\textwidth]{figures/loss/000000490171_synthesized_image_final.jpg}\\
(a) Input & (b) $\ell_2$ & (c) Ours & (d) Input & (e) $\ell_2$ & (f) Ours\\ 
\end{tabular} 
\caption{Effects of different types of reconstruction losses. Zoom in for best quality.}
\label{fig:ppl} 
\end{figure*}

\noindent\textbf{Comparison of Patch Perceptual Losses and $\ell_2$} We compare the quality of inpainting that is trained with Patch Perceptual Loss and $\ell_2$ loss. $\ell_2$ is used to train CE and GLI, but it does not correspond well to human perception of similarity. We investigate how our Patch Perceptual Loss compares with $\ell_2$ loss by training under the same condition but using different reconstruction losses (PPL vs $\ell_2$). From the test results, we see that our results are overwhelmingly better than $\ell_2$ results in terms of sharpness and coherence with context. Fig.~\ref{fig:ppl} shows two examples of comparison. 

\begin{figure*}[h!]  
\centering  
\small  
\setlength\tabcolsep{2pt}
\begin{tabular}{cc} 
\includegraphics[width=.45\textwidth]{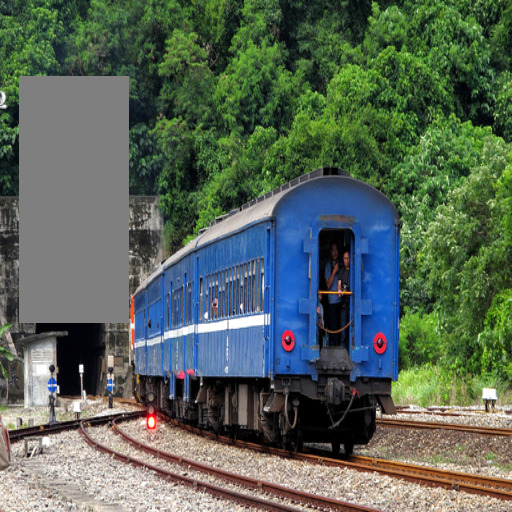}&
\includegraphics[width=.45\textwidth]{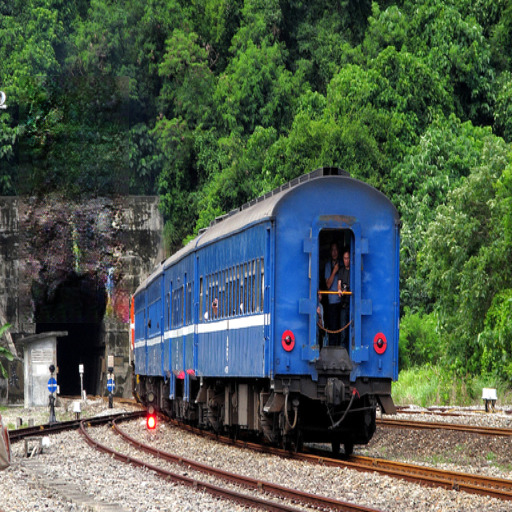}\\ 
\includegraphics[width=.45\textwidth]{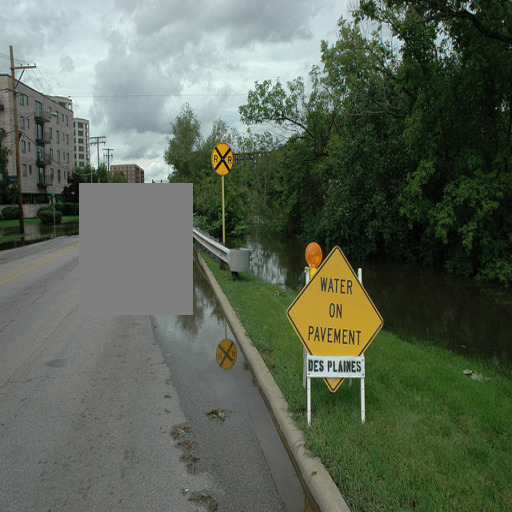}&
\includegraphics[width=.45\textwidth]{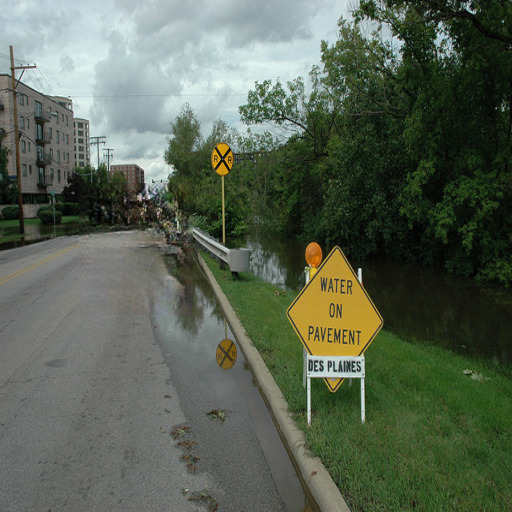}\\ 
(a) 512x512 Input & (b) Our Inpainting Result \\ 
\end{tabular} 
\caption{Inpainting results of 512x512 images.}
\label{fig:large} 
\end{figure*}  

\noindent\textbf{Results on Large Images} To showcase our ability to handle large images and holes, we show two inpainting results on 512x512 images in Fig.~\ref{fig:large}. Given the size of the receptive field and the depth of the network, our approach is able to inpaint visually plausible and semantically meaningful contents in these high-resolution images. 

\begin{figure*}[!ht]
\centering
\small
\setlength\tabcolsep{1pt}
\begin{tabular}{cccccc}
  \includegraphics[width=.17\textwidth]{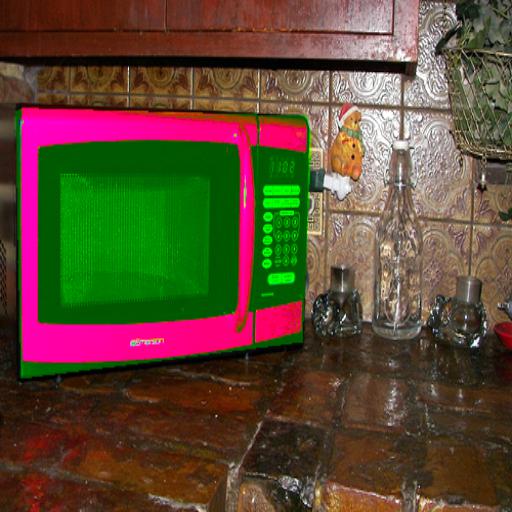}&
  \includegraphics[width=.17\textwidth]{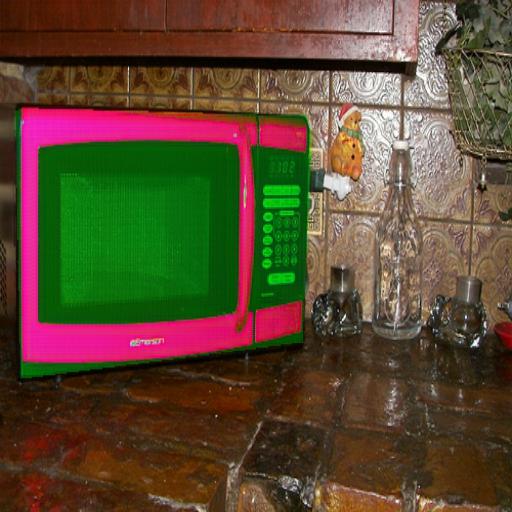}&
  \includegraphics[width=.17\textwidth]{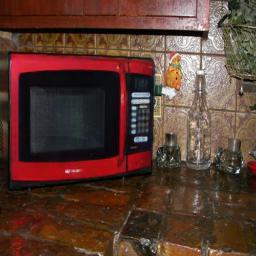}&
  \includegraphics[width=.17\textwidth]{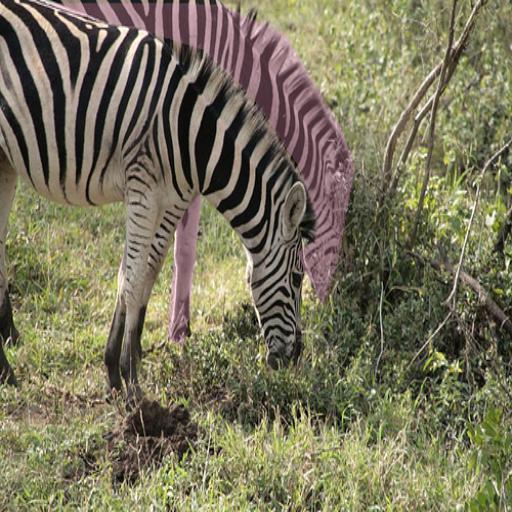}&
  \includegraphics[width=.17\textwidth]{figures/hm/000000159311_synthesized_image.jpg}&
  \includegraphics[width=.17\textwidth]{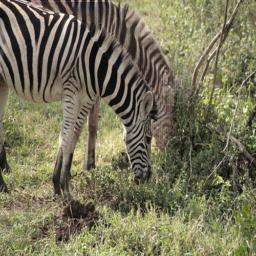} \\
  (a) Input & (b) DH~\cite{tsai2017deep}  & (c) Ours & (d) Input & (e) DH~\cite{tsai2017deep}  & (f) Ours \\
\end{tabular}
\caption{Examples of image harmonization results. For (a) and (d), the microwave and the zebra on the back have unusual color. Our method correctly adjusts their appearance and makes the images coherent and realistic. }
\label{fig:harm}
\end{figure*}

\begin{figure*}[!ht]
\centering
\small
\setlength\tabcolsep{1pt}
\begin{tabular}{cccc}
  \includegraphics[width=.25\textwidth]{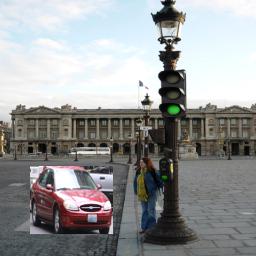}&
  \includegraphics[width=.25\textwidth]{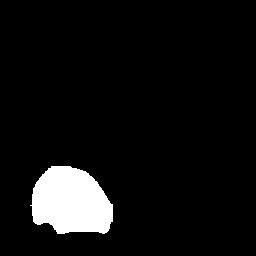}&
  \includegraphics[width=.25\textwidth]{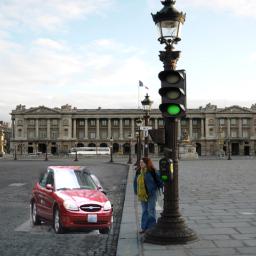}&
  \includegraphics[width=.25\textwidth]{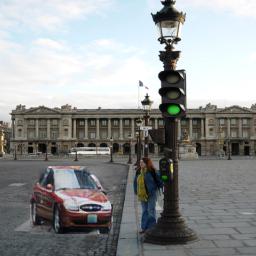} \\
  \includegraphics[width=.25\textwidth]{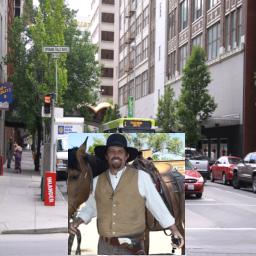}&
  \includegraphics[width=.25\textwidth]{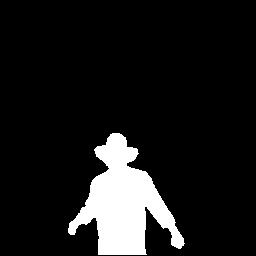}&
  \includegraphics[width=.25\textwidth]{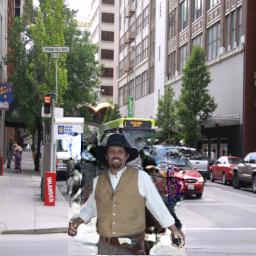}&
  \includegraphics[width=.25\textwidth]{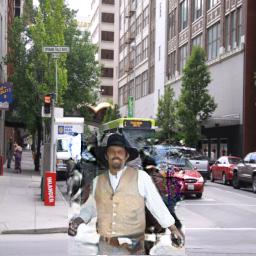} \\
  \includegraphics[width=.25\textwidth]{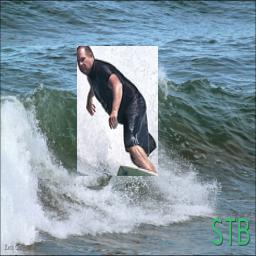}&
\includegraphics[width=.25\textwidth]{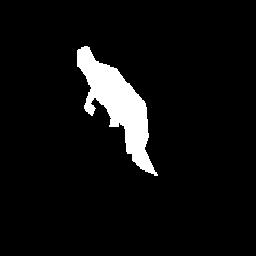}&
\includegraphics[width=.25\textwidth]{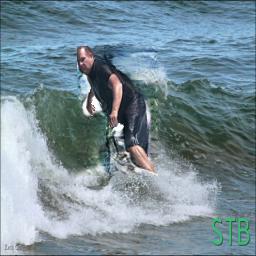}&
\includegraphics[width=.25\textwidth]{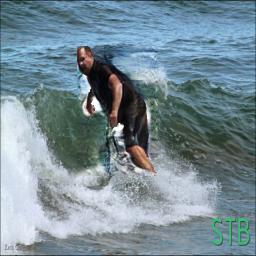}\\ 
\includegraphics[width=.25\textwidth]{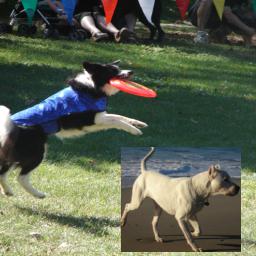}&
\includegraphics[width=.25\textwidth]{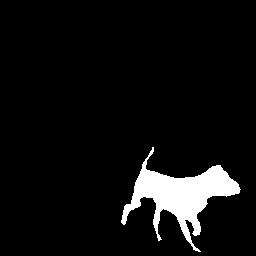}&
\includegraphics[width=.25\textwidth]{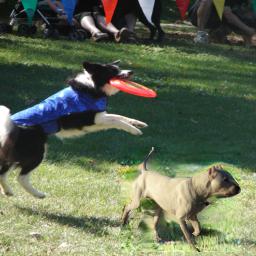}&
\includegraphics[width=.25\textwidth]{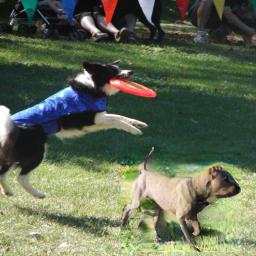}\\ 
  (a) Input & (b) Segmentation  & (c) Inpainting & (d) Final result  \\
\end{tabular}
\caption{Examples of interactive guided inpainting result. The segmentation mask is given by our foreground/background segmentation network trained on COCO. The final result combines the outputs of harmonization and inpainting.}
\label{fig:guided}
\end{figure*}

\subsection{Interactive Guided Inpainting}
\label{exp:guided} 

In this section, we consider object-based guided inpainting as a practical scenario. Specifically, we assume the user would like to add to the input image with objects from another guide image. Given an input image $I$ and a guide image $I_g$, we allow the user to select a region of $I_s$ by dragging a bounding box $B_g$ containing the object that he desires to add. Note that unlike previous settings of image composition, we do not require the user to accurately segment the object, but instead only providing a bounding box is sufficient. This greatly reduces the workload and simplifies the task. Next, we perform segmentation use a network to extract the object foreground. The segmentation network is adapted from Mask R-CNN~\cite{he2017mask} and trained on COCO, but is agnostic to object categories and only considers the foreground/background classification. Finally, we resize and paste $B_g$ to $I$. To make the composition look natural and realistic, we need to remove and inpaint the background of $B_g$ and also perform harmonization on the foreground object such that it is coherent the overall image appearance.   

To accomplish this task with known segmentation, we need to address two separate problems: inpainting and harmonization. Our model can be easily extended to train for both tasks at the same time, only requiring changing the input and output. To train this model, the data acquisition is similar to~\cite{tsai2017deep}, where we create artificially composited images using statistics color transfer~\cite{reinhard2001color} and another randomly chosen guide image. More specifically, at each iteration given the input image $I$, we randomly select another image $I_g$ from the dataset. We then select and segment an object $I_o$ from $I$ based on the segmentation mask of COCO. We then transfer the color from $I_g$ to $I_o$, and paste $I_o$ onto $I_g$ at a random location. Finally, we crop a bounding box $I_b$ from $I_g$ that contains $I_o$, and paste $I_b$ back to $I$. The result, together with the foreground/background segmentation mask of $I_b$, is given to the network as input. The model is modified to output two images, one for inpainting result and the other for harmonization result. We use the original $I$ as the ground truth for both tasks. 

The network jointly trained for inpainting and harmonization performs well in both tasks. Fig.~\ref{fig:harm} shows that it generates harmonization results better than~\cite{tsai2017deep}, which is the state-of-the-art deep harmonization method. Finally, in Fig.~\ref{fig:guided} we show several examples of interactive guided inpainting results.

%% file: 5_Conclusions.tex
\section{Conclusions}
We propose a novel model and training scheme to address the inpainting problem and achieve excellent performance on several tasks, including general inpainting, image harmonization and guided inpainting. Although we only explore the ability of our approach in image editing related tasks, we believe it is a general methodology that could be easily applied to other generative problems, such as image translation, image synthesis, etc. As future work, we plan to study the effectiveness of the proposed network and losses, and especially the procedural training scheme on a larger scope.